\documentclass[sigconf, 10pt]{acmart}
\renewcommand\footnotetextcopyrightpermission[1]{} % removes footnote with conference information in first column
\setcopyright{none}

\usepackage{algorithmic}
\usepackage{microtype}
\usepackage{graphicx}
\usepackage{booktabs} % for professional tables
\usepackage{caption}
\usepackage{subcaption}
\usepackage[utf8]{inputenc} % allow utf-8 input
\usepackage[T1]{fontenc}    % use 8-bit T1 fonts
\usepackage{hyperref}       % hyperlinks
\usepackage{url}            % simple URL typesetting
\usepackage{booktabs}       % professional-quality tables
\usepackage{amsfonts}       % blackboard math symbols
\usepackage{amsmath}
\usepackage{nicefrac}       % compact symbols for 1/2, etc.
\usepackage{microtype}      % microtypography
\usepackage{algorithm}
\usepackage{textcomp}
\usepackage{multirow}
\usepackage{numprint}
\usepackage{xcolor}
\usepackage{multirow}
\usepackage{enumitem}
\usepackage[group-separator={,}]{siunitx}
\npthousandsep{,}\npthousandthpartsep{}\npdecimalsign{.}
\usepackage{setspace}
\usepackage{hyperref}

\newcommand{\CPU}{{\tt CPU}}

\newcommand{\GNN}{{\tt GNN}}

\newcommand{\DL}{{\tt DL}}
\newcommand{\DGL}{{\tt DGL}}

\newcommand{\dgmb}{{DistGNN-MB}}
\newcommand{\he}{{\tt HE}}
\newcommand{\hec}{{\tt HEC}}
\newcommand{\AGG}{{\tt AGG}}
\newcommand{\UPDATE}{{\tt UPDATE}}

\newcommand{\VID}{{\tt VID}}

\begin{document}

\title[DistGNN-MB.]{DistGNN-MB: Distributed Large-Scale Graph Neural Network Training on x86 via Minibatch Sampling}

\author[Vasimuddin et. al.]{Vasimuddin Md, Ramanarayan Mohanty, Sanchit Misra, Sasikanth Avancha}
%\authornote{}
%\email{vasimuddin.md@intel.com}
%\author{Ramanarayan Mohanty}
%\email{vasimuddin.md@intel.com}
%\author{Sanchit Misra}
%\email{vasimuddin.md@intel.com}
%\author{Sasikanth Avancha}
\email{[vasimuddin.md, ramanarayan.mohanty, sanchit.misra, sasikanth.avancha]@intel.com}

%\email{webmaster@marysville-ohio.com}
\affiliation{%
  \institution{Intel Corporation}
  \streetaddress{}
  \city{}
  \state{}
  \country{}
  \postcode{}
}
\vskip 0.3in

\begin{abstract}
Training Graph Neural Networks, on graphs containing billions of vertices and edges, at scale using minibatch sampling poses a key challenge: strong-scaling graphs and training examples results in lower compute and higher communication volume and potential performance loss. DistGNN-MB employs a novel {\em Historical Embedding Cache} combined with compute-communication overlap to address this challenge. On a $32$-node ($64$-socket) cluster of 3rd generation Intel\textregistered \hspace{0.01cm} Xeon\textregistered \hspace{0.01cm} Scalable Processors with 36 cores per socket, DistGNN-MB trains $3$-layer GraphSAGE and GAT models on OGBN-Papers100M to convergence with epoch times of $2$ seconds and $4.9$ seconds, respectively, on $32$ compute nodes. At this scale, DistGNN-MB trains GraphSAGE $5.2\times$ faster than the widely-used DistDGL. 
DistGNN-MB trains GraphSAGE and GAT $10\times$ and $17.2\times$ faster, respectively, as compute nodes scale from $2$ to $32$.

%In an extreme-scale experiment, we observe that DistGNN-MB takes \revise{EP minutes} per epoch to train a web-graph with $3.5$B vertices and $28$B edges on $128$-node ($256$-socket) cluster.
\end{abstract}
%] % two column

\keywords{Graph Neural Networks, Minibatch Training, High Performance Computing, Machine Learning, Distributed Algorithms}

\maketitle

\section{Introduction}
\label{introduction}

Graph Neural Networks (GNN) are fast becoming a mainstream technology ingredient in applications such as recommendation systems~\cite{recommender}, fraud detection~\cite{fraud}, and large-scale drug discovery~\cite{zitnik2018modeling}. Interestingly, unlike Deep Learning techniques in vision or language modeling, GNN-based learning must adapt to the size and types of graphs whose structure the model must learn. Thus, for some applications, e.g., fraud detection, training the model on the full graph (aka full-batch training) will likely deliver better accuracy than sampling-based minibatch training, whereas for applications such as recommendation systems the latter is likely to be more scalable and accurate~\cite{recommender} . Additionally, the class of applications devoted to molecule search and discovery involves millions of "small" graphs each with only a few hundred or thousand vertices, requiring a combination of techniques employed in both full-graph and minibatch training to achieve good performance and high accuracy. This paper focuses on the {\em performance} of distributed, large-scale minibatch GNN training on CPUs.

In general, GNN training involves recursively executing (to the chosen depth) two key steps or primitives in order on chosen training examples: {\em Aggregation} (\AGG{}) and {\em \UPDATE{}}. In \AGG{}, source vertex or edge features are aggregated to destination vertices in the graph via {\em message passing} along the edges. In \UPDATE{}, aggregated features pass through a {\em Multi-Layer Perceptron} (MLP). Typical downstream tasks for a trained GNN model include node- (i.e., vertex), link- and graph-property prediction. Thus, the last hop of the \AGG{}-\UPDATE{} sequence consists of a classifier function to output predictions. A loss function, such as Cross-Entropy Loss or Negative-Log Likelihood compute the loss with respect to the target and errors backpropagate through \UPDATE{}-\AGG{} sequences. 

{\bf Minibatch GNN training} Similar to domains such as vision or language modeling, a GNN model uses a sampled minibatch of examples as inputs to execute \AGG{}-\UPDATE{} sequences for training. Unlike vision or language modeling, however, minibatch construction is more complex. GNN training examples are a set of vertices; each iteration constructs a {\em minibatch} of sub-graphs with a subset of the training vertices as roots. Each sub-graph is formed by recursively {\em sampling} a fixed number of neighbors starting from root vertices to the desired depth. Sampling can be uniform, biased or importance-based. The cost of minibatch creation can become a dominant component of overall training time; cost mitigation solutions include parallelization and asynchronous execution to overlap minibatch creation with compute. As discussed earlier, the \UPDATE{} primitive in GNN training typically consists of an MLP, which in turn, consists of a matrix-matrix multiplication ({\tt matmul}) operation followed by a non-linear function (e.g., ReLU or sigmoid) and a regularization function (e.g., Dropout). While {\tt matmul} is compute-intensive operation, ReLU and Dropout are memory-bandwidth intensive. Thus, the challenge is to ensure that these operations execute efficiently both with respect to the available compute power of the CPU as well as memory bandwidth.

{\bf Distributed GNN Training} GNNs naturally exhibit data parallelism because they are constructed either on instances of a single large graph (with billions of vertices and edges) or millions of small graphs (with only a few thousand vertices and edges). In the case of large-graphs, training GNN models requires partitioning them into smaller sub-graphs, and assigning each sub-graph to a CPU; it also requires splitting training example vertices in a balanced manner into the specified number of parts. Minimum-cut algorithms can partition such graphs along vertices~\cite{libra} or edges~\cite{karypis1997metis, karypis1997parmetis} into a specified number of sub-graphs. In the case of millions of small graphs (e.g., molecule graphs) training GNN models is an embarrassingly parallel problem, as each node in a cluster can feed a batch of example graphs in parallel to the model. 

{\bf Challenges of distributed, minibatch GNN training} Because we focus on training performance of large-scale GNNs in a distributed minibatch setting across CPUs via graph partitioning,  we consider the following. As a result of splitting training examples into smaller parts, with each part is assigned to a single CPU socket, the number of minibatches per socket reduces. Thus, not only does each CPU socket train the GNN model using a {\em graph partition}, but also samples fewer minibatches compared to single-CPU training. As a result of graph partitioning, \AGG{} now has two components: {\em local} and {\em remote}. The {\em performance} problem is now clear: as we partition graphs into smaller sub-graphs, each CPU socket uses fewer sub-graphs to train the model, and communication between CPU sockets to compute remote \AGG{} increases and begins to dominate at scale. The {\em accuracy} problem is that the size of the minibatch per socket remains the same; thus, the {\em global minibatch} scales with the number of CPU sockets potentially affecting accuracy. Thus, the constraint on improving performance via scaling  is to achieve the same (or within some $\epsilon$ of) target accuracy as single CPU training.  

%\paragraph{Contributions}
{\bf Contributions} This paper addresses these challenges and makes the following contributions:
\begin{itemize}
    \item A novel compute-communication overlap algorithm to reduce communication overhead at scale
    \item A novel, software-managed {\em Historical Embedding Cache} (\hec{}) data structure and associated algorithm that reduces communication without impacting accuracy
    \item Performance-optimized minibatch creation algorithm on CPUs to reduce overhead at scale
    \item Performance-optimized \UPDATE{} primitive on CPUs to ensure efficient execution
    \item Demonstrating large-scale GNN minibatch training using the open-source, widely-used Deep Graph Library (DGL) on general-purpose, CPU-only clusters with high performance 
\end{itemize}

The rest of the paper is organized as follows: Section~\ref{background} discusses GNN preliminaries along with the GraphSAGE~\cite{hamilton2017inductive} and Graph Attention Network (GAT)~\cite{velickovic2018graph} models. In section~\ref{methods}, we detail our Asynchronous Embedding Push {\tt AEP} algorithm which uses \hec{}, along with performance optimizations to minibatch sampling, \AGG{} and \UPDATE{} primitives. Section~\ref{results} presents and discusses performance and accuracy evaluation results in detail. In section~\ref{related} we briefly describe related research in large-scale GNN minibatch training. We present concluding remarks and discuss future work in section~\ref{conclusions}.
\section{Background}
\label{background}

Given an input graph $\mathcal{G}(\mathcal{V}, \mathcal{E}), (v, \mathcal{N}(v)) \in \mathcal{V}$ (where $\mathcal{N}$($v$) is the neighborhood of $v$), and an $\mathcal{L}$-layer GNN constructed on $\mathcal{G}$, GraphSAGE (shown in equation~\ref{sage-eqn}) performs \AGG{} and \UPDATE{} for each layer $l$ on vertex features $f^l$. In distributed training, $\mathcal{N}(v)$ may reside on the local CPU or on a remote CPU. Thus, \AGG{} is executed as {\em local} and {\em remote} (section~\ref{sec-remote-agg}) operations, with the latter resulting in the need for communications. GraphSAGE consists of two {\tt matmul} operations: one transforms neighbor aggregates ${h^l}_{\mathcal{N}(v)}$ using model parameters ${W^l}_n$ (where $n$ denotes "neighborhood") and the other transforms the vertex's own features ${h^l}_v$ using model parameters ${W^l}_s$ (where $s$ denotes "self"). GraphSAGE uses ReLU to non-linearize embeddings and Dropout to generate the final, regularized output ${h^l}_v$ for layer $l$. In distributed training, CPUs perform {\tt all-reduce} communication to exchange model parameter gradients and update model parameters locally.
%\begin{flalign*} 
%&{h^l}_{\mathcal{N}(v)} = \AGG{}^l(\left\{f^{l-1}_u | u \in \mathcal{N}(v) \right\})
%\end{flalign*}
% &{h^l}_v = Drop(ReLU({W^l}_n \cdot {h^l}_{\mathcal{N}(v)} + {W^l}_m \cdot {h^l}_v + b^l)
%\end{flalign*}
\begin{equation} \label{sage-eqn}
\begin{split}
&{h^l}_{\mathcal{N}(v)} = \AGG{}^l(\left\{f^{l-1}_u | u \in \mathcal{N}(v) \right\})\\
&{h^l}_v = {\tt Dropout}({\tt ReLU}({W^l}_n \cdot {h^l}_{\mathcal{N}(v)} + {W^l}_s \cdot {h^l}_v + b^l)
\end{split}
\end{equation}
Similarly, on an input graph $\mathcal{G}$ and a $\mathcal{L}$-layer GNN, the GAT model (shown in equation~\ref{gat-eqn}) performs \AGG{} and \UPDATE{} for each layer $l$ on vertex features. In this work, we modify DGL's implementation of GAT to enable performance optimizations; we have empirically determined that these modifications do not materially impact accuracy. Our modification applies bias and non-linearity to the output of dot-product between embeddings and model parameters {\em before} computing attention co-efficients ${\alpha^l}_{uv}$. The dot-product, edge-feature and attention-coefficient computation operations execute locally on each machine. \AGG{} executes in a distributed manner as described above.
\begin{equation} \label{gat-eqn}
\begin{split}
&{z^l}_u = {\tt ReLU}({W^l}_u \cdot {f^l}_u + {b^l})\\
&{z^l}_v = {\tt ReLU}({W^l}_v \cdot {f^l}_v + {b^l})\\
&{e^l}_u = {W^l}_{au} \circ {z^l}_u\\
&{e^l}_v = {W^l}_{av} \circ {z^l}_v\\
&{\alpha^l}_{uv} = {\tt EdgeSoftmax}({\tt LeakyRELU}({e^l}_u + {e^l}_v))\\
&{h^{l+1}}_v = \AGG{}^l({{\alpha}^l}_{uv} {z^l}_u | u \in \mathcal{N}(v)))
\end{split}
\end{equation}
\section{Distributed Minibatch Training}
\label{methods}

GNN training is demanding in terms of compute resources, i.e., CPUs, memory capacity and memory bandwidth. Training massive graphs on a single CPU socket is challenging due to such resource requirements. \AGG{} is a memory-intensive operation, with a byte-to-op ratio $\gg 1$; e.g., aggregating two $256$-element {\tt FP}$32$ tensors with $+$ operator requires $3072$ bytes to be Read/Written, with only $256$ compute operations -- a byte-to-op ratio of $12$. Thus, using naive \AGG{} and \UPDATE{} implementations for large-scale GNN training on a single CPU socket will result in poor performance.

In this section, we discuss algorithms for distributed, large-scale minibatch GNN training including important single-socket CPU optimizations; these algorithms and optimizations enable DisGNN-MB to achieve at least an order-of-magnitude speedup over naive single-socket CPU implementations. 

The key to GNN model training in a distributed environment is partitioning the underlying graph into smaller sub-graphs (section~\ref{partitioning}). Because \AGG{} contains a {\em reduction} operator, graph partitioning will automatically result in feature vector (or vertex embedding) communication to ensure correct and complete the \AGG{} operation. To reduce communication overhead, the graph partitioning algorithm usually cuts minimum number of edges or vertices while creating partitions with balanced training examples. In this work, we use a modified version of the popular  algorithm, Metis, which employs a minimum-edge-cut technique to partition graphs.

However, while balanced min-edge-cut partitioning is necessary for communication-reduction, it is not sufficient to hide communication-cost. An un-optimized \AGG{} implementation may potentially create a large performance bottleneck by communicating vertex embeddings for every layer in each minibatch, if these communications cannot be overlapped with compute. Common techniques to mitigate this problem fall into two broad approaches: communication volume {\em reduction} per iteration and {\em avoidance} across iterations. 

In this paper, we describe a suite of algorithms and a data structure called \hec{} for volume reduction via {\em communication delay}; to further reduce overall cost, we {\em overlap} communications with local compute on every CPU socket in the distributed system. (For ease of usage, we refer to CPU sockets in a distributed system as {\em ranks} in the rest of the paper.)

\subsection{Metis Graph Partitioning with Balance}
\label{partitioning}
In addition to minimal edge-cuts, GNNs have a specific requirement: training vertices must be distributed among graph partitions to minimize load imbalance across ranks. DistDGL~\cite{distdgl} augments Metis meet this requirement. We define key terms required to denote various entities as a result of min-edge-cut partitioning: {\em solid} and {\em halo} vertices, original ($\VID{}_o$), partition ($\VID{}_p$) and batch vertex IDs ($\VID{}_b$), respectively. 

Let an edge $e = (u, v)$ in the full graph $\mathcal{G}$ be a min-cut edge during partitioning. Thus, $e$ is cut into two edges ${e^{\prime}} = (u, v^{\prime})$ and $e^{\prime\prime} = (u^{\prime}, v)$. We refer to $u^{\prime}$ and $v^{\prime}$ as {\em halo} vertices, which represent their corresponding {\em solid} vertices $u$ and $v$ in the other partition, respectively. Halo vertices $u^{\prime}$ and $v^{\prime}$ do not contain input features. During \AGG{}, solid vertices $u$ and $v$ communicate embeddings to $u^{\prime}$ and $v^{\prime}$, respectively; these embeddings are aggregated with their corresponding destination vertices via message passing within their respective partitions. 
Per definition of $\mathcal{G}$, $\VID{}_o = range(0...\mathcal{V}-1)$. Let the number of vertices in a partition be $\mathcal{P}$; then $\VID{}_p = range(0...\mathcal{P}-1)$. Finally, let the number of vertices in a minibatch be $\mathcal{B}$; then $\VID{}_b = range(0...\mathcal{B}-1)$. DistGNN-MB uses lookup tables to maintain correspondence between $\VID{}_o$, $\VID{}_p$, and $\VID{}_b$.

%To ensure consecutive IDs for all nodes within a partition, (which we refer to as {\em partition nodes}), DGL-Metis re-numbers them starting from $0$; we refer to these node IDs as PIDs without loss of generality.  
%To perform communication, the original node ID (OID) is required to uniquely identify a node among the partitions.
%Each partition node stores the corresponding original node ID (OID), which uniquely identifies local PIDs, required for communication.
%Additionally, minibatch creation via node sampling within a partition results in their re-numbering from $0$ in each instance; we refer to such nodes as {\em batch nodes} and associated IDs as batch IDs (BID). Multiple levels of PIDs require conversion among different types of node IDs during minibatch training. We use lookup tables for mapping among node ID types.

% Partitioning
% Node numbering
% MBC
% Training: Local Agg, remote agg, update
% Local agg and update are exactly same as single socket
% single socket optimizations
% Remote AGG

\begin{figure*}[!ht]
\begin{subfigure}{0.48\textwidth}
    \centering
        \includegraphics[width=\linewidth]{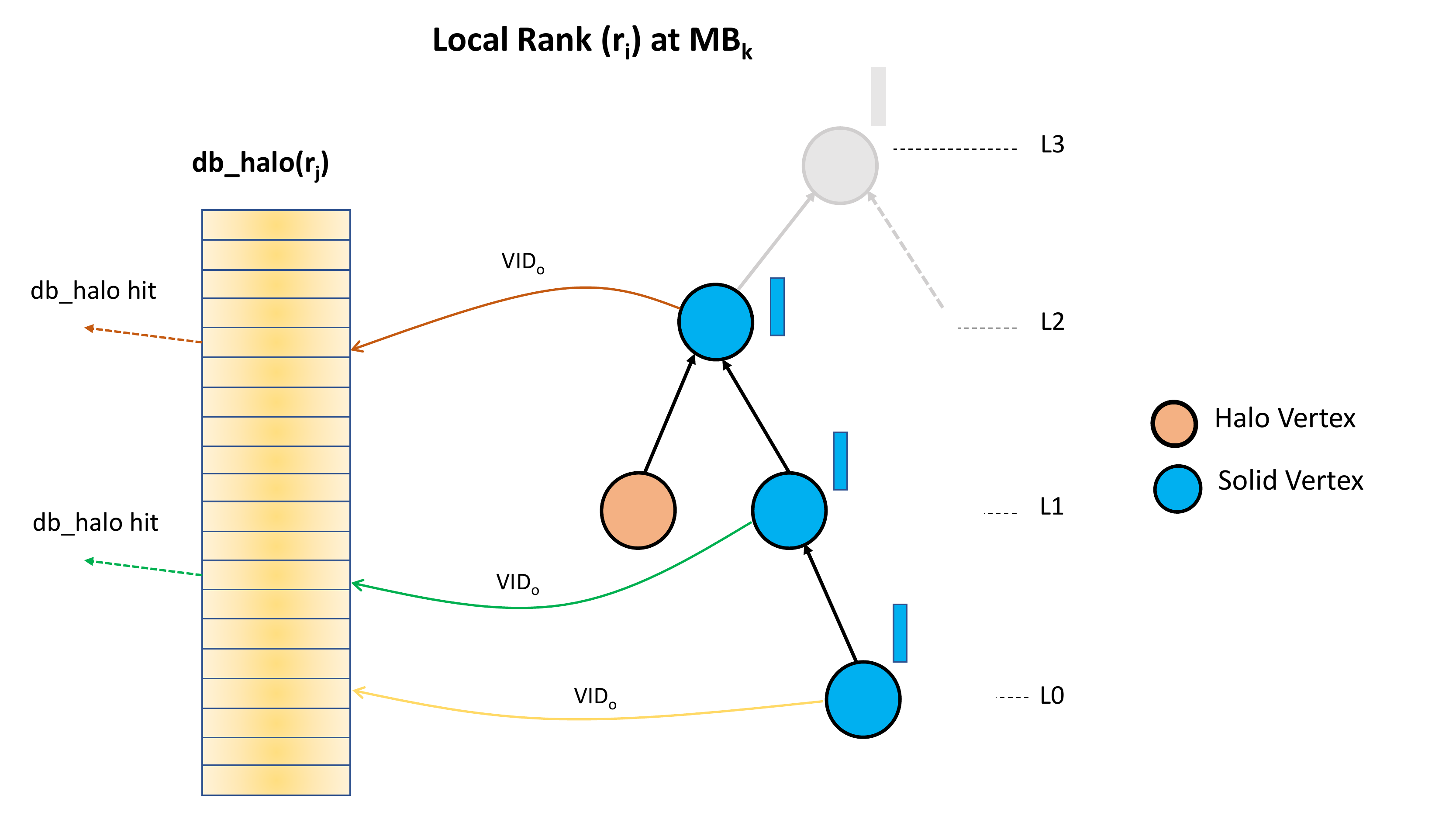}
    \caption{}
    \label{}
\end{subfigure}
\begin{subfigure}{0.48\textwidth}
    \centering
        \includegraphics[width=\linewidth]{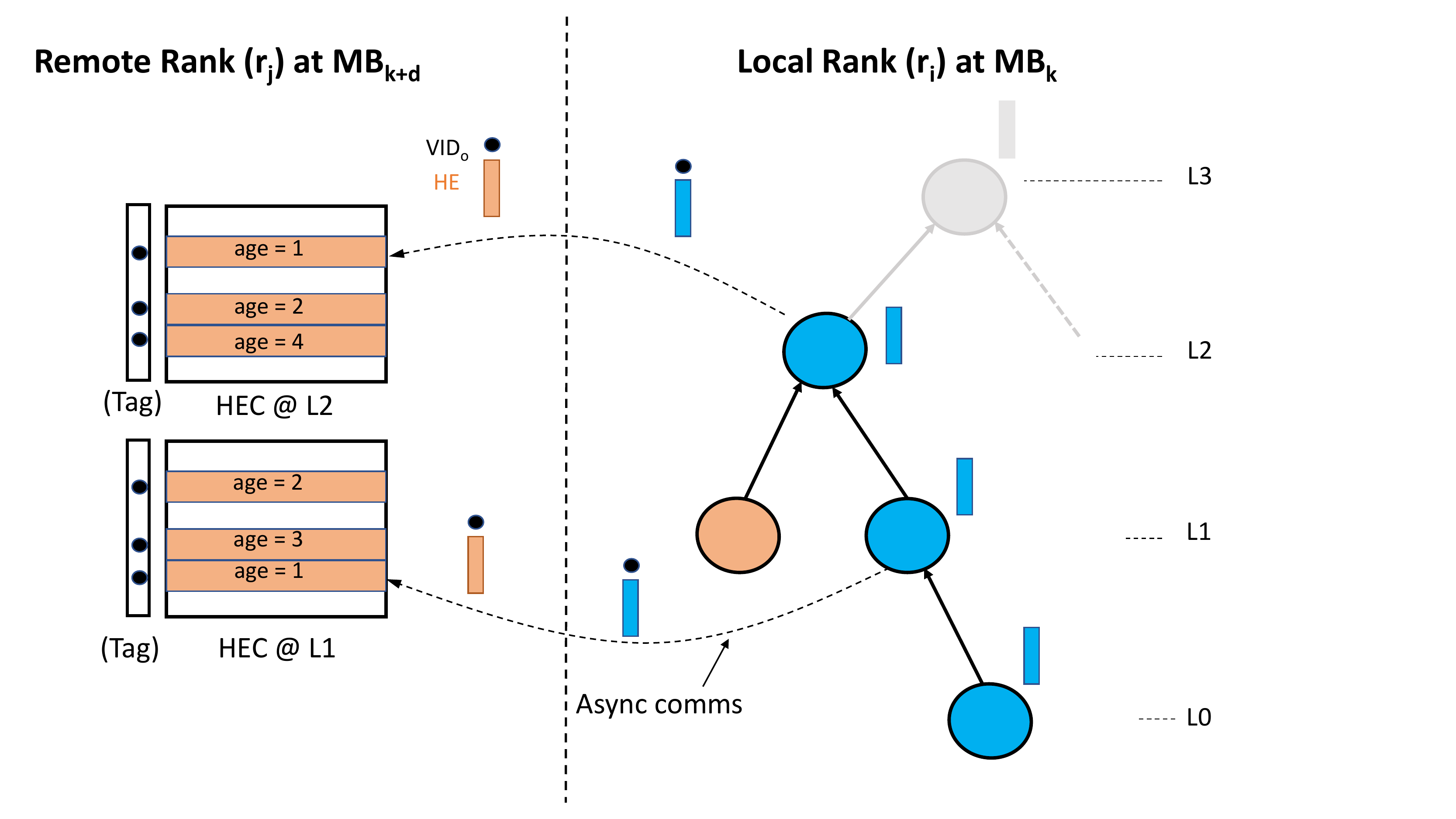}
    \caption{}
    \label{}
\end{subfigure}
\begin{subfigure}{0.48\textwidth}
    \centering
        \includegraphics[width=\linewidth]{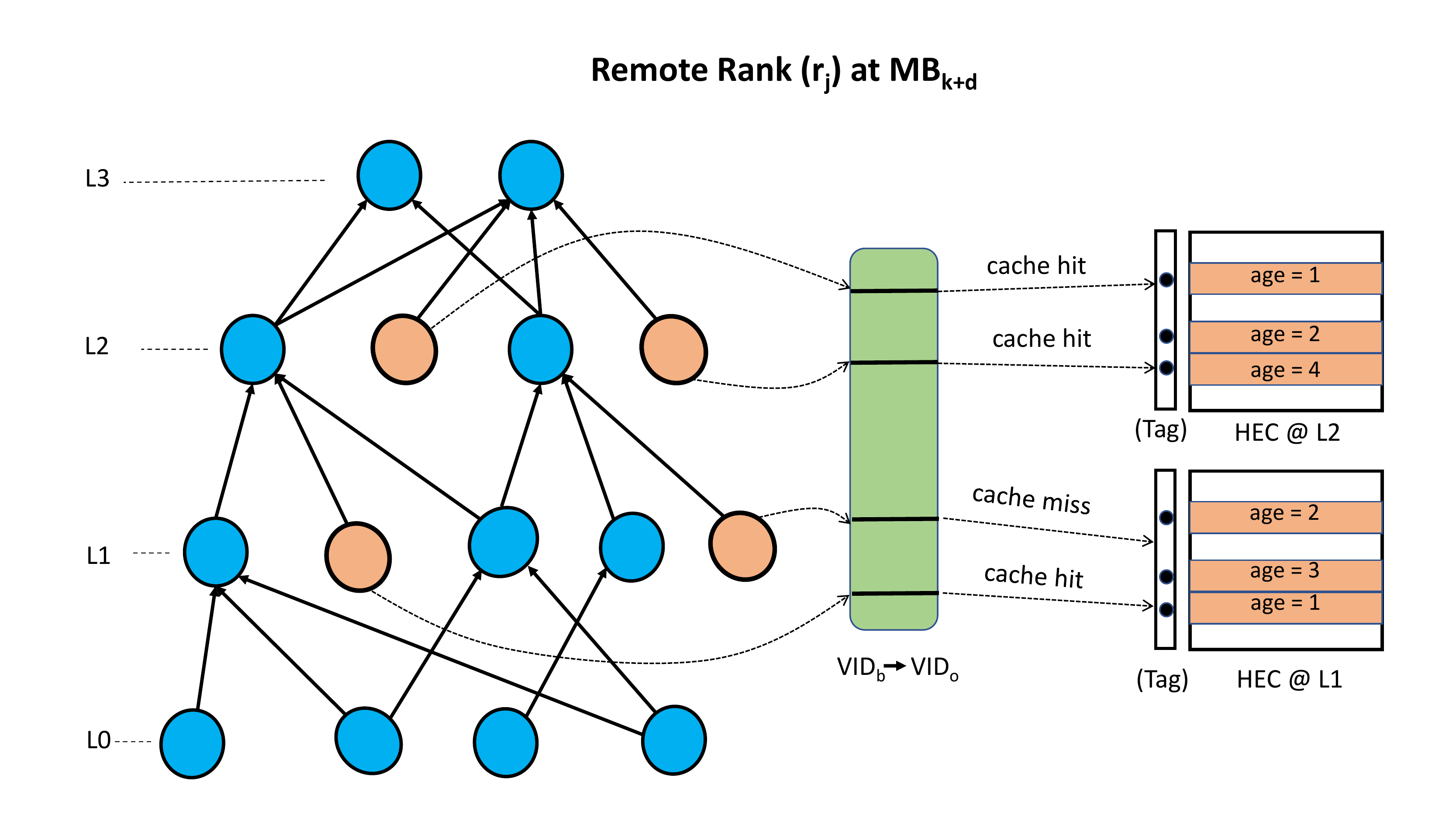}
    \caption{}
    \label{}
\end{subfigure}
\begin{subfigure}{0.48\textwidth}
    \centering
        \includegraphics[width=\linewidth]{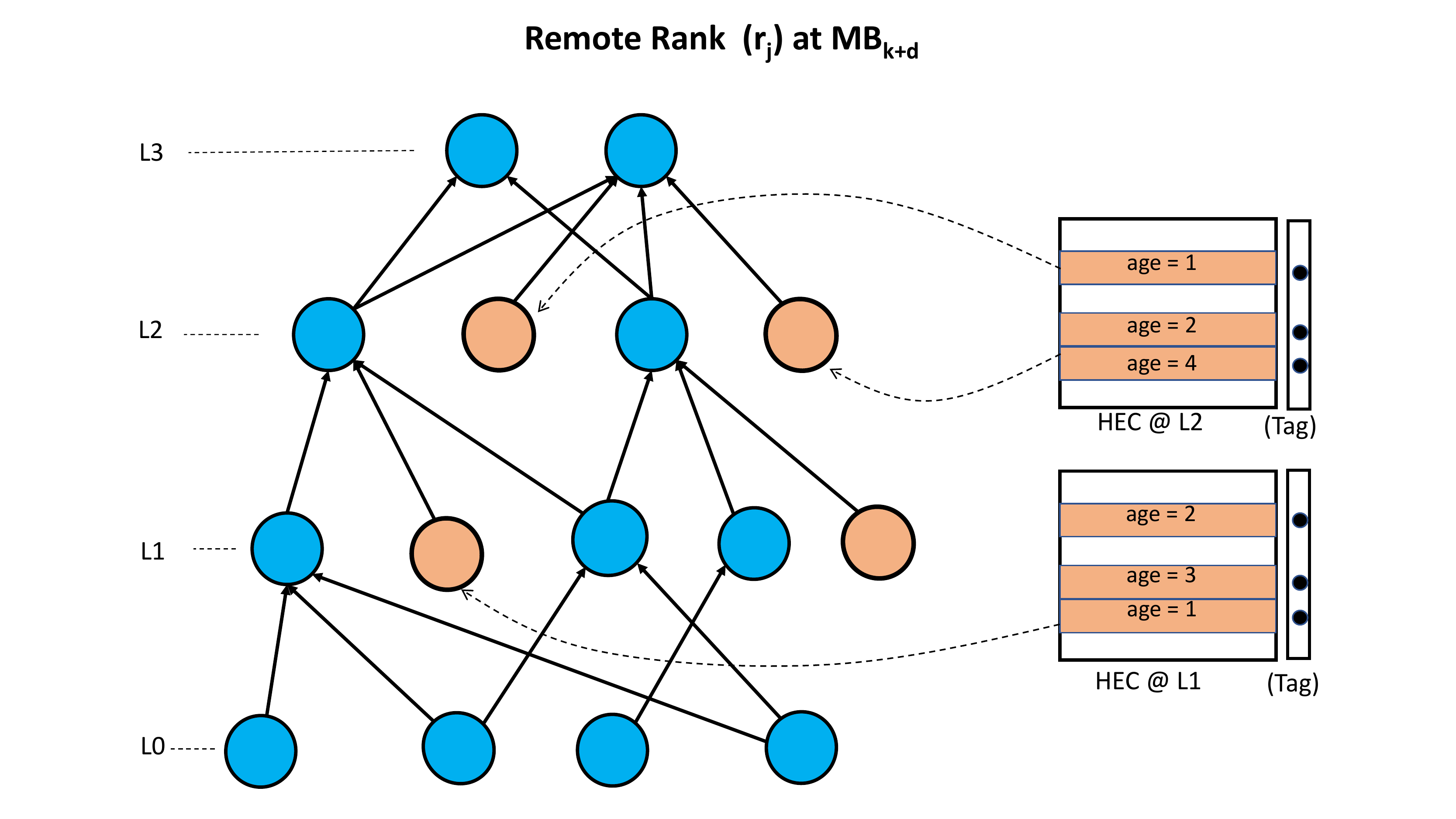}
    \caption{}
    \label{}
\end{subfigure}
\vspace{-10pt}
\caption{$n$ layers in minibatch (MB) are labelled as L$_0$, L$_1$ \ldots L$_{n-1}$, with L$_{n-1}$ containing training seed vertices. Minibatch contains solid and halo vertices. Solid-vertex embeddings communication is asynchronous, with delay $d$, to fill remote \hec{}s. Halo-vertex search occurs in local \hec{} and corresponding embeddings are used during MB \AGG{}. Each inner MB layer maintains an \hec{}. (a) Each remote rank $r_j$  searches for solid vertices of local MB in {\tt db\_halo}; embdeddings of vertices with {\tt db\_halo} hit become eligible for communication to remote \hec{}s. (b) rank $r_i$, processing $k^{th}$ minibatch, MB$_k$, asynchronously communicates the selected solid vertices ($\VID{}_o$) and their embeddings to \hec{} of remote rank $r_j$; $r_j$ receives the communication while processing of $(k+d)^{th}$ minibatch, MB$_{k+d}$, overlapping communication with computation of $d$ minibatches. Received embeddings are cached in local \hec{} with $age$ set for the occupied cache-lines. (c) Before performing \AGG{}, rank $r_j$ searches for its halo vertices in local \hec{}. Accessing local \hec{} requires conversion from $\VID{}_b$ to $\VID{}_o$ as $\VID{}_o$ are maintained as \hec{} tags. (d) Embeddings for halo vertices with a cache-hit are loaded from \hec{} during \AGG{}.}
\label{fig:mb-hist-emb}
\end{figure*}

%\subsection{Asynchronous Push-based Communication Model}
\subsection{Remote Aggregation using Delayed, Historical Embeddings}
\label{sec-remote-agg}
Local aggregation executes as shown in equations~\ref{sage-eqn} and ~\ref{gat-eqn} in section~\ref{background}. In this section, we focus exclusively on optimized remote aggregation to reduce communication cost. 

To accomplish {\em remote} aggregation, we employ the Historical Embedding (\he{}) concept. As~\cite{chen2017stochastic} describe, \he{}s are embeddings from past training iterations, which under conditions of {\em bounded staleness}~\cite{fey2021gnnautoscale},~\cite{sancus} contribute usefully to model convergence (and more importantly, do not degrade accuracy). The key advantage of \he{} in GNN training is in saving embedding communication time by using a locally available, albeit {\em stale} version to complete the \AGG{} step. To improve \he{} utility and reuse, we introduce \hec{}, a software-managed cache containing \he{} as cache-lines. Each rank in the distributed system creates and associates an \hec{} with each GNN layer. \hec{} has a fixed size $cs$ and each cache-line has a maximum life-span $ls$; all cache-lines with age greater than $ls$ are purged from \hec{}. $\VID{}_o$ serve as tags to quickly find and access cache-lines in \hec{}. Cache-line replacement in \hec{} follows oldest cache-line first ({\tt OCF}) policy. This ensures fresher embeddings in the \hec{}.

\hec{} plays a key role in ensuring compute-communication overlap during remote aggregation by serving as a repository for {\em delayed embeddings}, as shown in figure~\ref{fig:mb-hist-emb}(b); e.g., a local rank generates an embedding \he{} at MB$_k$ that a remote rank receives at MB$_{k+d}$, after a delay of $d$ iterations, which \hec{} stores (either by replacing an existing cache-line with a matching tag, replacing an expired cache-line or filling a new cache line entry). An additional benefit of \hec{} is that unlike DistDGL, ranks do not need to prefetch vertices from remote neighbors after minibatch creation to determine connectivity at run-time and establish "remote edges" for message passing; they merely reuse cached \he{} via $\VID{}_o$ tags received and stored as part of communication processing.

\hec{} management consists of three key operations: {\tt HECSearch}, {\tt HECLoad} and {\tt HECStore}. For each halo vertex $hv$, {\tt HECSearch} searches \hec{} efficiently for resident \he{} $f_{hv}$ using $\VID{}_o$ as the corresponding tag and returns a pointer to the matching line on a hit. {\tt HECLoad} gathers embeddings $f$ using memory pointers (that {\tt HECSearch} returns) and stores them into the minibatch data structure for \AGG{}. {\tt HECStore} scatters embeddings received from remote partitions via asynchronous communication into the \hec{}. We have optimized these management functions to perform lookup, gather and scatter operations efficiently using OpenMP\textregistered\ parallel regions.

%\subsubsection{Communication Data-structure Management}
One of the most important data structures in DistGNN-MB is {\tt db\_halo}. On each rank in the distributed system, it stores original vertex IDs $\VID{}_o$ associated with {\em halo} vertices from remote partitions (whose {\em solid} avatars exist local partition). Each minibatch sampled locally contains both {\em solid} and {\em halo} vertices. Using mappings in {\tt db\_halo}, our optimized communication algorithm retrieves features/embeddings required by remote {\em halo} vertices $\VID{}_o$ and asynchronously communicates them to remote partitions. The {\tt Map} function to map solid to halo vertices during minibatch training is one of the most expensive operations in DistGNN-MB and we optimize it using OpenMP\textregistered\ parallel regions.

The graph partition data-structure that Metis creates is also very important. It maintains details about each graph vertex in a lookup table ({\tt LUT}). These details include markers to identify vertex type (solid or halo) and $\VID{}_o$. In DGL, the minibatch creation process provides the LUT to map minibatch vertex IDs $\VID{}_b$ to partition IDs $\VID{}_p$. Partition IDs can be used to index into the graph LUT to identify the corresponding vertex type ({\tt findHaloNodes()} and {\tt findSolidNodes()} in Algorithm~\ref{algo:push-model}) and original vertex IDs $\VID{}_o$ (for \hec{} access and communication). We optimize these functions also using OpenMP parallel regions.

% \begin{figure*}
%     \centering
%     \includegraphics[width=\linewidth]{figures/he.pdf}
%     \caption{Use of historical embeddings in minibatch}
%     \label{fig:mb-hist-emb}
% \end{figure*}

\subsubsection{Asynchronous Embedding Push (AEP) Algorithm}
%% motivation
% Communication of embeddings for \AGG{} is an expensive operation, in terms of both communication latency and pre- and post-processing for the communication.
% Such execution time cost exacerbates as we scale the number of ranks. In this work, we attempt to minimize the communication latency cost by overlapping communication with computation. We leverage \hec{} to manage the overlap.   

% push model description
We apply the {\tt AEP} algorithm to fill \hec{} with historical embeddings, optimize remote aggregation and reduce communication overhead during distributed minibatch GNN training. At each GNN layer, {\tt AEP} {\em pushes} solid-vertex embeddings from the current local minibatch to remote \hec{}s, asynchronously. {\tt AEP} overlaps communication across iterations, with minibatch computation and communication-data-structure management functions. Remote \hec{}s fill with \he{} cache-lines in preparation for their use in subsequent local minibatches. 

%In section~\ref{results}, we showcase healthy \hec{} hit-rate under various cache parameter settings.

In line $2$ of Algorithm~\ref{algo:push-model}, all ranks initialize their \hec{} (as described in Algorithm~\ref{algo:init}), at each GNN layer with parameters $cs$, $nc$ and $ls$ described earlier. 
Additionally, each rank is assigned a graph partition $P$. Each rank creates its {\tt db\_halo} database using remote partition $hv$, as shown in Algorithm~\ref{algo:init}.
 
Lines $3$-$27$ run for a configurable number of epochs. After creating minibatches (MB) and extracting seeds on Line $4$, the algorithm iterates through all $M$ minibatches on Line $5$. Line $6$ gathers minibatch features using seeds as indices. 

On Lines $7$-$13$, each rank does the following: On Line $8$, it determines whether to wait for communication of delayed remote partition halo vertices ($h_v$) and corresponding embeddings ($f_{hv}$) (figure~\ref{fig:mb-hist-emb}(b)). If so, and upon receipt, each rank stores $f_{hv}$ into its \hec{} on Line $9$. On Lines $10$-$11$ (figure~\ref{fig:mb-hist-emb}(c)), each rank extracts minibatch halo vertices and uses them to search through its \hec{} for embeddings. It combines them with minibatch embeddings ($f_{mb}$) (figure~\ref{fig:mb-hist-emb}(d)). On Line $11$, a rank may encounter a cache miss in its \hec{} -- if so, it eliminates the corresponding $hv$ from minibatch execution. Now, \AGG{} in equations~\ref{sage-eqn} and~\ref{gat-eqn} can execute on $f_{mb}$, thus completing the overall \AGG{} operation. As part of training, this is followed by \UPDATE{}.  

Each rank executes Lines $15$-$24$ to asynchronously communicate solid-vertex embeddings to remote ranks if minibatch iteration $k$ is within $M-d$ iterations. On Line $15$, each rank extracts its minibatch solid vertices and uses them to retrieve a subset $sv^{\prime}$ from {\tt db\_halo} (Line 18) as shown in figure~\ref{fig:mb-hist-emb}(a). Each rank restricts the number of retrieved solid vertices to a configurable parameter $nc$; it does this by randomly sampling a subset of $sv^{\prime}$ based on their degree (Line 20). On Line 22, each rank gathers features ($f_{sv^{\prime}}$) associated with $sv^{\prime}$ locally. Finally, on Line 24, all ranks send $f_{sv^{\prime}}$ to remote ranks (figure~\ref{fig:mb-hist-emb}(b)). Execution repeats on Line 8 and continues until all epochs have completed.

\begin{algorithm}[h]
   \caption{Initialize} 
   \label{algo:init}
\begin{algorithmic}[1]
\small
\REQUIRE Graph partition $P$
\REQUIRE \hec{}
\REQUIRE \hec{} parameters: $cs$, $nc$, $ls$
\REQUIRE Partition halo vertices, $hv$
\STATE Assign $P$ to a rank
\STATE $B$ $\leftarrow$ {\tt Bcast}($hv$)
\STATE {\tt db\_halo} $\leftarrow{}$ {\tt CreateDB}($B$)
\STATE $\hec{}$.cs = cs
\STATE $\hec{}$.nc = nc
%\STATE $\hec{}$.d = d
\STATE $\hec{}$.ls = ls
\end{algorithmic}
\end{algorithm}

\begin{algorithm}[h]
   \caption{Asynchronous Embedding Push} 
   \label{algo:push-model}
\begin{algorithmic}[1]
\small
\REQUIRE Per partition vertex features, $f$
\REQUIRE Per partition training vertices, $tv$
\REQUIRE Per partition halo vertices, $hv$
\REQUIRE Per partition solid vertices, $sv$
\REQUIRE Per rank communication object, $C$
\REQUIRE Communication delay, $d$
\FOR{each rank in parallel}
\STATE Initialize()   \hspace{.5in}\COMMENT{Initialize \hec{} and {\tt db\_halo}}
\FOR {{\tt epoch}}
%\STATE Get $tv$ to $seed$ $M$ minibatches ($mb$)
\STATE $seeds$, $M$ $\leftarrow{}$  {\tt CreateMinibatches}($tv$)  
\FOR{each $mb$, $0 \le k < M$ (at each layer)} 
    \STATE $f_{mb}$ $\leftarrow{}$ {\tt LoadMBFeatures}($seeds$, $f$)
    \IF{$k \ge d$}
      \STATE $hv, f_{hv}$ $\leftarrow{}$ {\tt comm\_wait($C$)}   \label{wait}
      \STATE {\tt HECStore}($hv$, $f_{hv}$)    \label{hecstore}
      \STATE  $hv$ $\leftarrow{}$  {\tt findHaloNodes}($mb$) 
      \STATE $hv^{\prime}$ $\leftarrow{}$ {\tt HECSearch}($hv$)   \label{hecsearch}
      \STATE $f_{mb}$ $\leftarrow{}$ $f_{mb}\ \cup \ $ {\tt HECLoad}($hv^{\prime}$)     \label{hecload}
    \ENDIF
    \IF{$k < M-d$}
      \STATE $sv$ $\leftarrow{}$ {\tt findSolidNodes}($mb$) 
      \STATE $f_{sv} \leftarrow{} 0 $
      \FOR{each remote rank}
        \STATE $sv^{\prime}$ $\leftarrow{}$ {\tt Map}($sv$, {\tt db\_halo})   \label{map}
        \IF{(|$sv^{\prime}$|) > $\hec{}.nc$}
          \STATE  $sv^{\prime}$ $\leftarrow{}$ {\tt Sample}($sv^{\prime}$, {\tt deg}($sv^{\prime}$))    \label{sample}
        \ENDIF
        %\STATE $f_u$ $\leftarrow{}$ {\tt gather}(${sv^{\prime}}$)  
        \STATE $f_{sv^{\prime}}$ $\leftarrow{}$  $f_{sv^{\prime}} \cup \ {\tt gather}({sv^{\prime}}) $
 \label{gather}
      \ENDFOR
      \STATE $C$ $\leftarrow{}$  {\tt AlltoallAsync}($f_{sv^{\prime}}$)  \label{async}
    \ENDIF
    %\STATE Train(model, mb, fmb)
    %\STATE Alltoall() $\rightarrow$ model\_gradients
    %\STATE update\_model(model\_gradients)
\ENDFOR
\ENDFOR
\ENDFOR				
\end{algorithmic}
\end{algorithm}

\subsection{Single-socket CPU Optimizations}
{\bf Thread-Parallel Minibatch Sampling} In general, minibatch sampling is a very expensive operation and naive implementations can easily result in huge overheads, dominating overall execution time. Unlike DGL and other approaches that use {\em distributed} minibatch sampling, we choose to sample locally and use historical embeddings along with limited communication to achieve accuracy. Thus, instead of using a number of samplers to overlap minibatch sampling with computation, we implement it as a thread-parallel, {\em synchronous} operation. We observe that, both on a single-socket CPU and at scale, this implementation results in low overhead minibatch sampling with respect to the overall training time. 

{\bf Broadcast Support for AGG} 
% Given the large variety of binary/unary and reduction operators that different GNN algorithms can use in \AGG{}, applying SIMD to \AGG{} using manually written intrinsics could be a time consuming task.
The LIBXSMM library ~\cite{libxsmm} provides highly architecture
optimized primitives for many matrix operations including our use-cases. It uses JITing to generate optimal assembly code with SIMD intrinsics where applicable, thus providing
more instruction reduction than manually written intrinsics based code. 
DGL already uses LIBXSMM for several such use-cases. One use-case that is missing is when each value of one or both the input feature vectors has to be broadcasted multiple times to form a new input feature vector that is used in \AGG{}. For example, in GAT, the input feature vector corresponds to one attention head and has to be replicated as many times as there are attention heads. DGL uses a simple scalar loop for this use-case. In our implementation, we have made the necessary modifications in LIBXSMM and DGL to add support for SIMD for this use-case.

{\bf UPDATE Optimizations} We implement optimized versions of the \UPDATE{} parts of GraphSAGE and GAT equations discussed in section~\ref{background} in a separate PyTorch C++ extension and use LIBXSMM for this purpose as well. We discuss these optimizations next.

{\bf Operator Fusion} We optimize the \UPDATE{} operation -- in the second equation in GraphSAGE and the first four equations in GAT (described in section~\ref{background}) -- via operator-fusion, blocking, thread-parallelization and LIBXSMM TPPs. Given that each outer operator uses the output of the next inner operator as input (e.g., ReLU uses the output of the dot-product, and Dropout uses that of ReLU), we optimize the code path by {\em fusing} all these operators together, which results in re-using inner operator output (present in the L2 cache of each CPU core) as next outer operator input. This approach saves main-memory bandwith and improves training time significantly. We also fuse operations within the inner-most computation in the referred equations (e.g., adding bias $b^l$ to the dot-product output).

\begin{table*}[t]
\caption{\GNN{} benchmark datasets. Directed edges in the original graph are converted to un-directed edges.}
\centering
%\vspace{-10pt}
\begin{tabular}{l|r|r|r|r|r | r}
\hline
Datasets	       &  \#vertex	              &  \#edge	                                  & \#feat            &	\#class            &  \#train vertices   &  \#test vertices   \\ \hline 
OGBN-Products    &	$\numprint{2449029}$    &	$\numprint{123718280}$                    & $\numprint{100}$	& $\numprint{47}$    &     $\numprint{196615}$ & $\numprint{2213091}$  \\   
OGBN-Papers100M  &	$\numprint{111059956}$	& $\numprint{3231371744}$  & $\numprint{128}$  &	$\numprint{172}$   &    $\numprint{1207179}$  & $\numprint{214338}$   \\
%% MAG240M-LSC      &	$\numprint{244160499}$	& $\textcolor{red}{\numprint{3456728464}}$  & $\numprint{768}$  &	$\numprint{153}$   &     1 &  1    \\
%Web-Graph        &	$\numprint{3438368599}$	& $\textcolor{red}{\numprint{56120691584}}$ & $\numprint{50}$   &	$\numprint{272}$   &    1  & 1   \\
\hline
\end{tabular}
\label{gnn-datasets}
\end{table*}

{\bf Blocking} To ensure that the output tensors from inner operations remain in the L2 cache, we transform input and weight tensors to the dot-product operation from 2-D to 4-D in memory, i.e., $in[N][C]$ becomes $in[nn][bn][nc][bc]$ and $wt[K][C]$ becomes $wt[nk][nc][bc][bk]$ (where $bn$, $bc$, $bk$ are block-sizes that ensure no L2-bank conflicts and optimal register-file usage). Thus, with the $nc$ dimension as an outer-loop, the dot-product operation executes on small matrices $in[bn][bc]$ and $wt[bc][bk]$ to produce a small matrix $out[bn][bk]$ in every iteration. We perform remainder handling in cases where $N$ is not divisible by $bn$; if $C$ or $K$ are not divisible by $bc$ or $bk$, respectively, then $bc = C$ and $bk = K$. 

{\bf Thread-Parallelization} We launch OpenMP regions with the number of threads in the region equal to the number of cores per CPU socket. We map the outermost tensor dimension $nn$ of the output tensor to OpenMP threads, thus speeding-up traversal over its iteration space. In neural network training, weight gradient computation (i.e., Backward-by-Weight ({\tt BWD\_W})) becomes very expensive if the minibatch dimension $N$ is much larger (1-3 orders of magnitude) compared to weight dimensions $C$ and $K$. GNN training exhibits exactly this pattern, where $N$ can be $400K$, $50K$ or $5K$, whereas weight dimensions are $100$, $128$ or $256$ in both GraphSAGE and GAT models. Thus, for {\tt BWD\_W}, we parallelize along the $N$ dimension, let each thread compute its own copy of the weight gradients and subsequently perform a reduction across all copies to arrive at the final weight gradient.

%% Enumerate the kernels here

%\subsubsection*{Data Loader Reconfiguration:}

%\subsubsection*{MLP Optimizations:}
%% Characterize the kernels (fwd, bck) here: Matmul, Dropout, Attention etc.
%% Describe the individual optimizations and their reasoning

%\subsubsection*{SpMM Optimizations:}
%% SpMM bcast

\section{Experimental Evaluation}
\label{results}

\subsection{Experiment Setup}
\label{sec-results-exp-setup}

We run single-socket experiments on 3rd generation Intel\textregistered \hspace{0.01cm} Xeon\textregistered \hspace{0.01cm} $8380$ \CPU{} @$2.30$ {\tt GHz} with $40$ cores (single socket), equipped
with $256$ {\tt GB} of memory per socket, running CentOS $8.1$; the theoretical peak bandwidth to DRAM on this machine is $204$ {\tt GB}/s. For distributed memory runs, we use a compute cluster with 3rd generation Intel\textregistered \hspace{0.01cm} Xeon\textregistered \hspace{0.01cm} $8360Y$ \CPU{} @$2.40$ GHz with $36$ cores per socket in a dual-socket system and $128$ {\tt GB} memory, running CentOS $8.1$. These compute nodes use a Mellanox HDR interconnect with DragonFly topology for inter-node communications.

We use GCC v$8.5.0$ to compile \DGL{} and our PyTorch C++ extension. 
We use \DGL{}v$0.8.x$ with PyTorch v1.10.2 as the backend \DL{} framework to demonstrate performance of our solutions. 
We use PyTorch Autograd profiler to profile application performance. For communication in distributed setting, we use the MPI (version 2021.4.0.3347) implementation in the oneAPI library.

In all our distributed experiments, we run $1$ rank per socket.
We average GraphSAGE performance measurements over $20$ and $40$ epochs for OGBN-Products and OGBN-Papers100M, respectively; and for GAT over $20$ and $50$ epochs for OGBN-Products and OGBN-Papers100M  respectively.

\subsection{Datasets and Models}
\label{sec-dataset}
Table~\ref{gnn-datasets} shows the details of the two \GNN{} benchmark datasets: OGBN-Products, OGBN-Papers100M, used in our experiments.
OGBN-Products and OGBN-Papers100M are part of the Open Graph Benchmark (OGB)~\cite{hu2020ogb} designed to measure Node Property Prediction accuracy. In the rest of the paper, we refer to them as OGBN datasets.

%\textcolor{red}{Todo: Web-Graph description}

Table~\ref{tab:model-params} shows hyperparameter settings for GraphSAGE and GAT. Fan-out fixes the number of neighbors sampled at each layer. GraphSAGE applies {\tt mean} operator for neighborhood aggregation, while GAT applies {\tt gcn} operator. We set the minibatch size to $1000$ for all our experiments.

%\vspace{-10pt}
\begin{table}[htb]
\caption{GraphSAGE and GAT parameters for OGBN datasets.}
\centering
%\vspace{-10pt}
\begin{tabular}{l|c|c}
\hline
Parameters         & GraphSAGE  & GAT      \\  \hline
fan-out            &  5,10,15   & 5,10,15                  \\
%batch size         &  1000      &  1000                     \\
aggregator          &  {\it mean}      &  {\it gcn}                      \\
hidden size        &  256       &  256                      \\
\#hidden layers    &  2        & 2                           \\
\#attn. heads            &  -        & 4                          \\
Single-socket CPU lr     &  $0.003$        &  $0.001$                         \\
Multi-socket CPU lr      &  $0.006$        &   $0.001$                        \\
\hline
\end{tabular}
\label{tab:model-params}
\end{table}

To enable distributed training, we employ data parallelism paradigm. In data parallelism, each parallel process (MPI rank) maintains an exact copy of the model, while the input graph is partitioned and each rank processes one partition. Consequently, MPI ranks execute a blocking { All-Reduce} communication operation to exchange and reduce model parameter gradients after every iteration to update local model parameters.

%% move this as one section in multinode
\subsection{Single-socket CPU Performance for \dgmb{}}
In this section, we present performance improvements to DGL due to architecture-aware single-socket CPU optimizations discussed in section~\ref{methods}. 

\begin{figure*}[!ht]
\centering
\begin{minipage}{0.75\linewidth}
{
	\subfloat[GraphSAGE on OGBN-Products]
	{
		\includegraphics[width=0.38\linewidth]{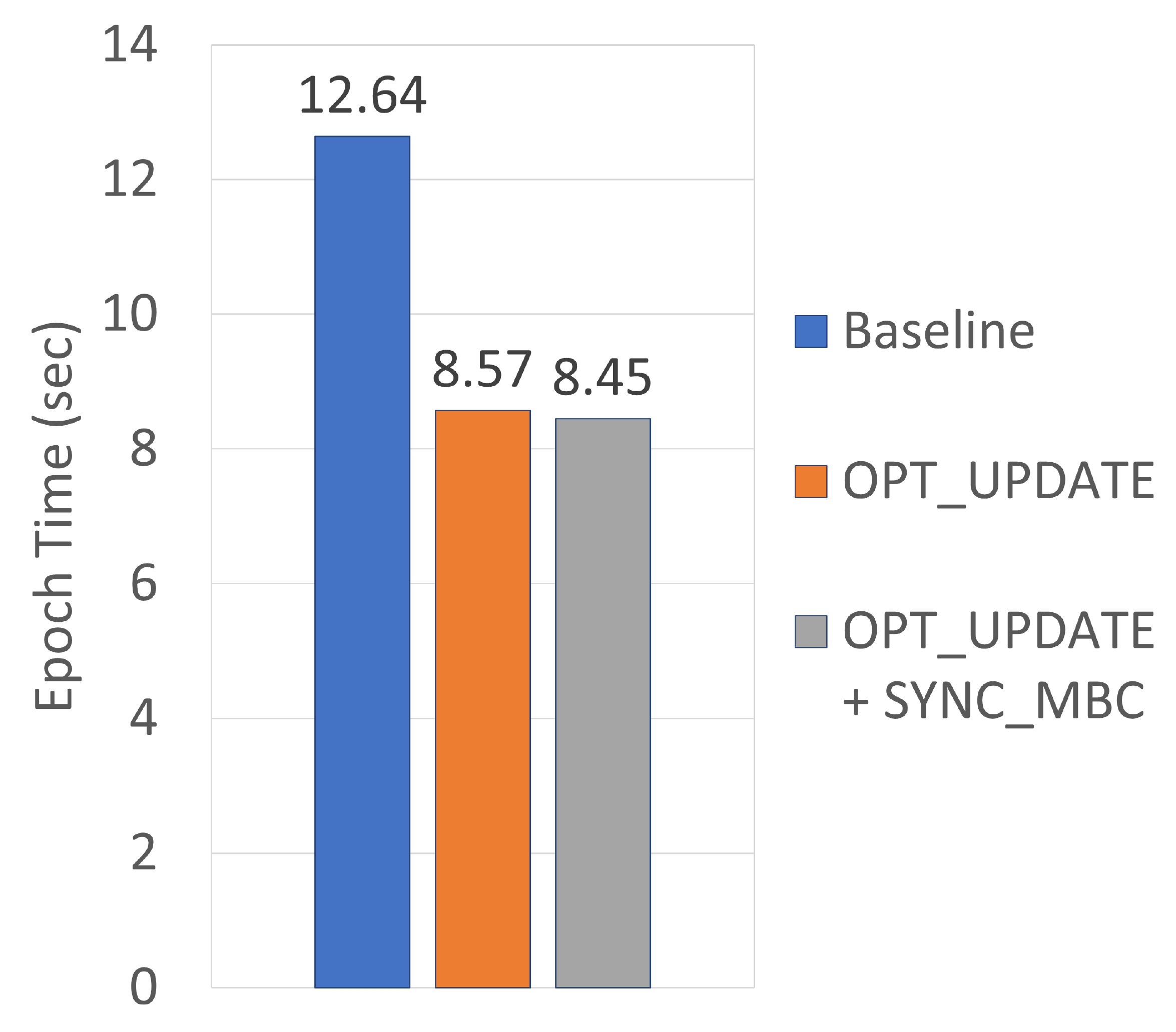}
            \label{fig:graphsageperfss_prod}
	}
	\hfill
	\subfloat[GraphSAGE on OGBN-Papers100M]
	{
		\includegraphics[width=0.38\linewidth]{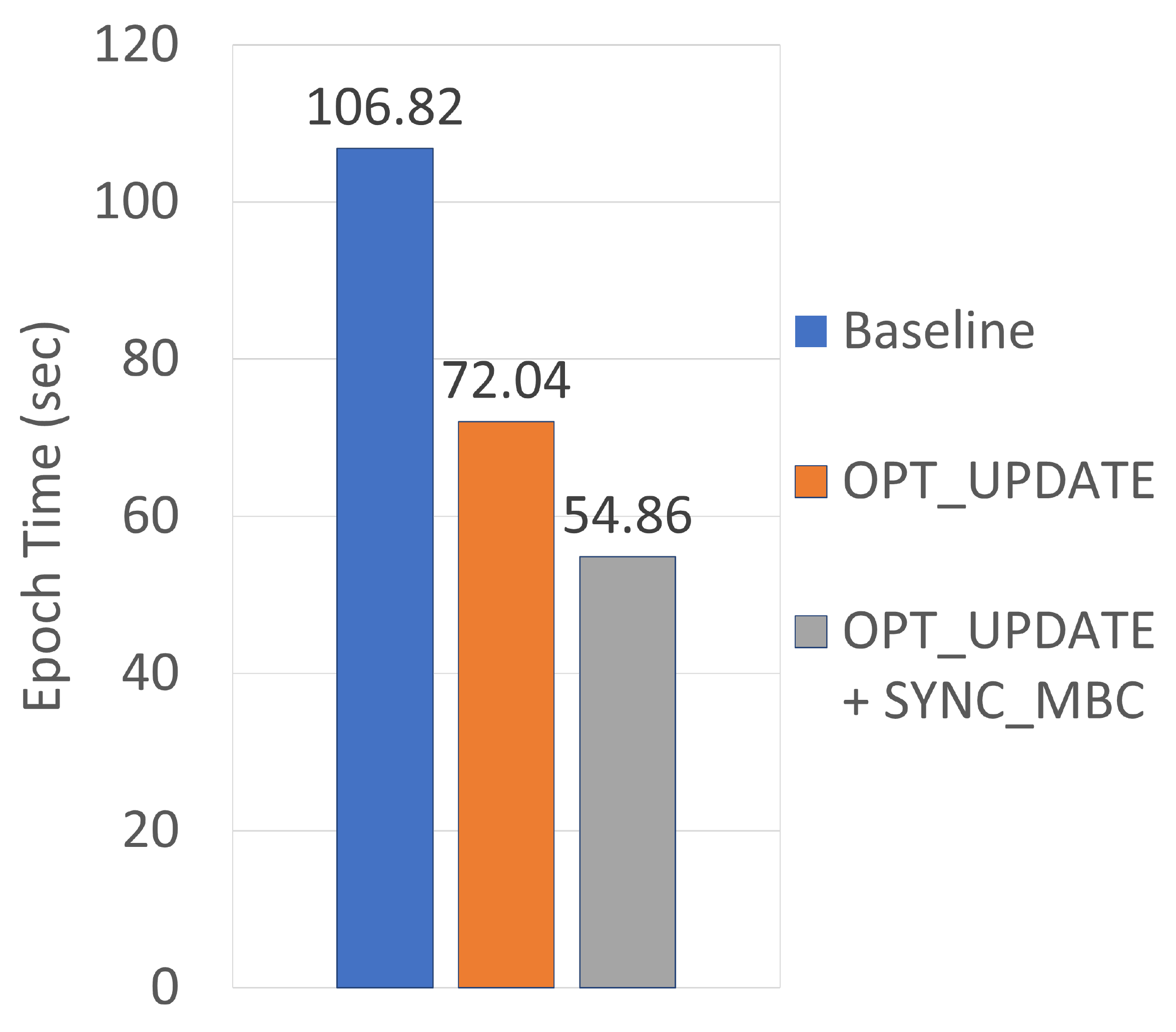}
            \label{fig:graphsageperfss_paper}
            
	}
	\vfill
	\subfloat[GAT on OGBN-Products]
	{
	    \includegraphics[width=0.38\linewidth]{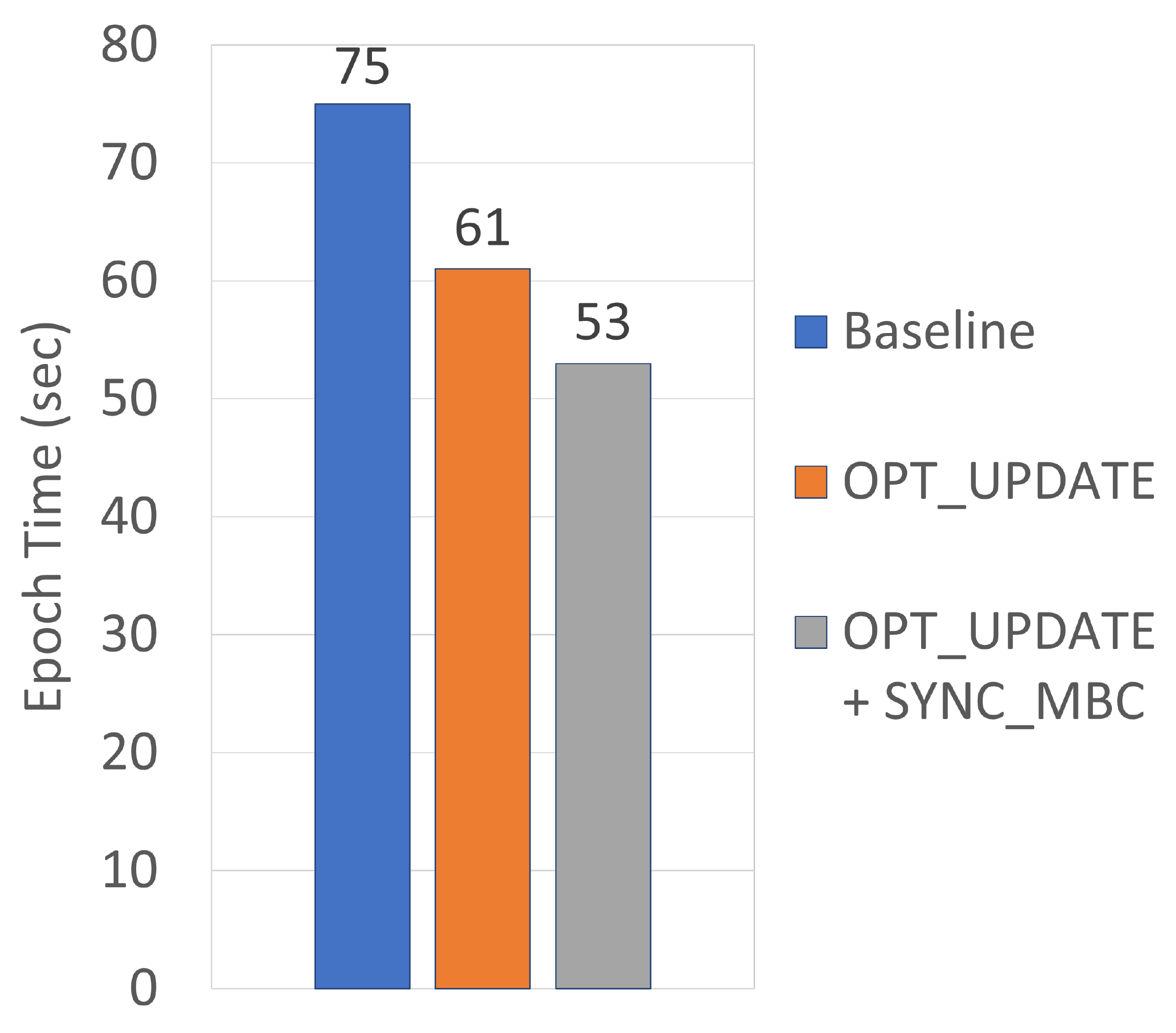}
            \label{fig:gatperfss_prod}
	}
	\hfill
	\subfloat[GAT on OGBN-Papers100M]
	{
	    \includegraphics[width=0.38\linewidth]{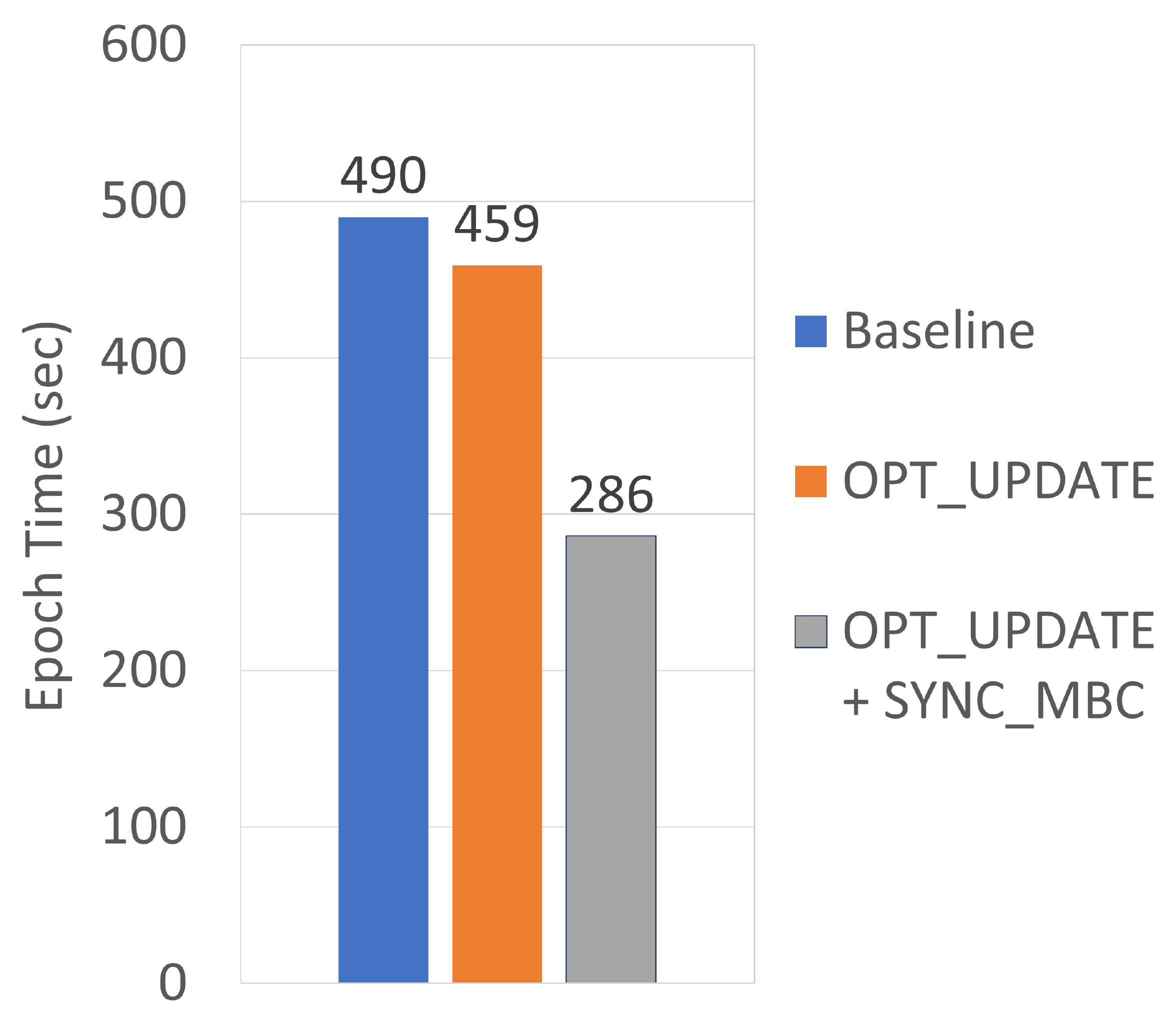}
            \label{fig:gatperfss_paper}
	}
}
\end{minipage}
\vspace{-10pt}
\caption{Comparison of single-socket CPU epoch time of DGL baseline with our optimized GraphSAGE and GAT models using OGBN datasets. We use batch size of $1000$. {\tt Baseline} is the original DGL execution time. {\tt OPT\_UPDATE} is the execution time with our \UPDATE{} optimizations. {\tt OPT\_UPDATE + SYNC\_MBC} is the execution time with parallel synchronized minibatch sampler. {\tt OPT\_UPDATE} for GAT incorporates optimized Broadcast operation. Labels at the top of the bars are epoch time.}
\label{fig:ss-perf}
\end{figure*}

We refer to the original DGL code as baseline. {\tt OPT\_UPDATE} in the figure~\ref{fig:ss-perf} refers to the execution time of our optimized {\tt UPDATE} operation. Similarly, {\tt OPT\_UPDATE + SYNC\_MBC} refers to execution time improvement due to synchronized parallel minibatch sampler on the top of {\tt OPT\_UPDATE} improvements. 
Figure~\ref{fig:ss-perf} shows single-socket CPU performance.
All our optimizations render GraphSAGE $1.5\times$ and $2\times$ faster on OGBN-Products and OGBN-Papers100M, respectively, compared to the baseline. Optimized {\tt UPDATE} gains $44\%$ and $48\%$ execution time improvements over baseline on OGBN-Products and OGBN-Papers100M, respectively.

Similarly, all our optimizations results in GAT to be $1.4\times$ and $1.7\times$ faster on OGBN-Products and OGBN-Papers100M, respectively, over the baseline. 
Optimizations to {\tt Broadcast} component of \AGG{} that is used in GAT speed it up by $2.22\times$ and $1.69\times$ for OGBN-Products and OGBN-Papers100M, respectively, over the baseline. 
The combination of optimized {\tt UPDATE} and Broadcast results in $23\%$ and $6\%$ execution time improvements over the baseline on OGBN-Products and OGBN-Papers100M, respectively.
% This results in acceleration of \AGG{} by $14\%$ and $36\%$ for GAT on OGBN-Products and OGBN-Papers100M, respectively. 

\subsection{Multi-Socket Performance for \dgmb{}}
% GraphSAGE: ogbn-products, ogbn-papers100M, web-graph
% GAT: ogbn-products, ogbn-papers100M
%% ranks 4,8,16,32,64
%% Time: ep, dltime, FWD, BWD, AReduce, optimizer
%% Scaling: ep, dltime, FWD/total_agg/local_agg, BWD
%% imbalance, remote agg time

In this section, we discuss performance results, including execution time and scaling of our distributed algorithms, and discuss key factors impacting their efficiency. 
%Table~\ref{tab:model-params} dictates various \he{} parameter settings for our experiments. 
We maintain the following \hec{} parameter settings throughout our experiments: HEC size ($cs$) is $1$M entries, Cache-line communication threshold ($nc$) is $2000$, Cache-line life-span ($ls$) is $2$, and Communication Delay ($d$) is $1$ iteration.

\begin{figure*}[!ht]
\begin{subfigure}{0.48\textwidth}    
    \centering
    \includegraphics[width=\textwidth]{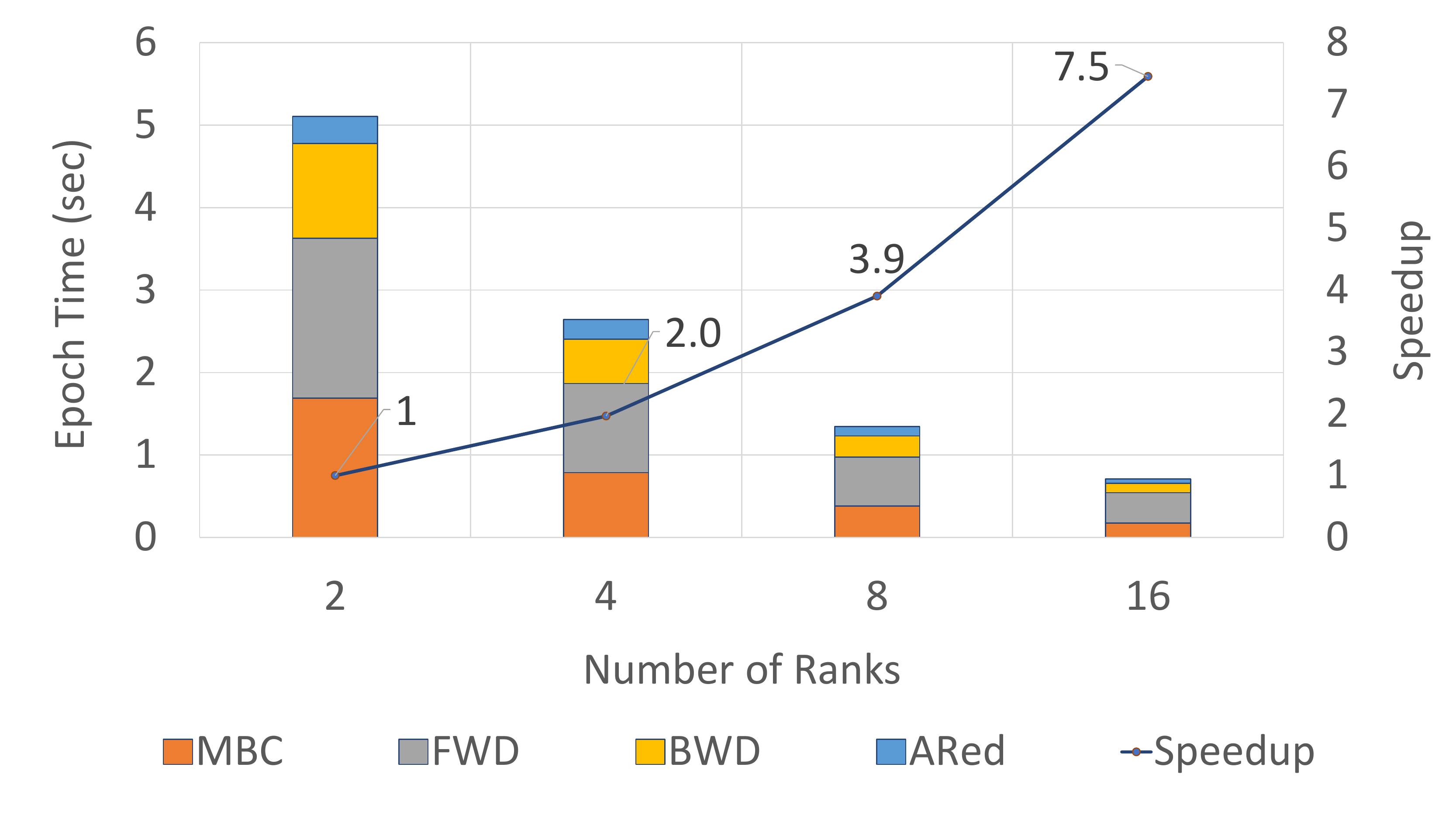}
    \caption*{OGBN-Products}
    \label{fig:gsage-perf-1}    
\end{subfigure}
\begin{subfigure}{0.48\textwidth}
    \centering
    \includegraphics[width=\textwidth]{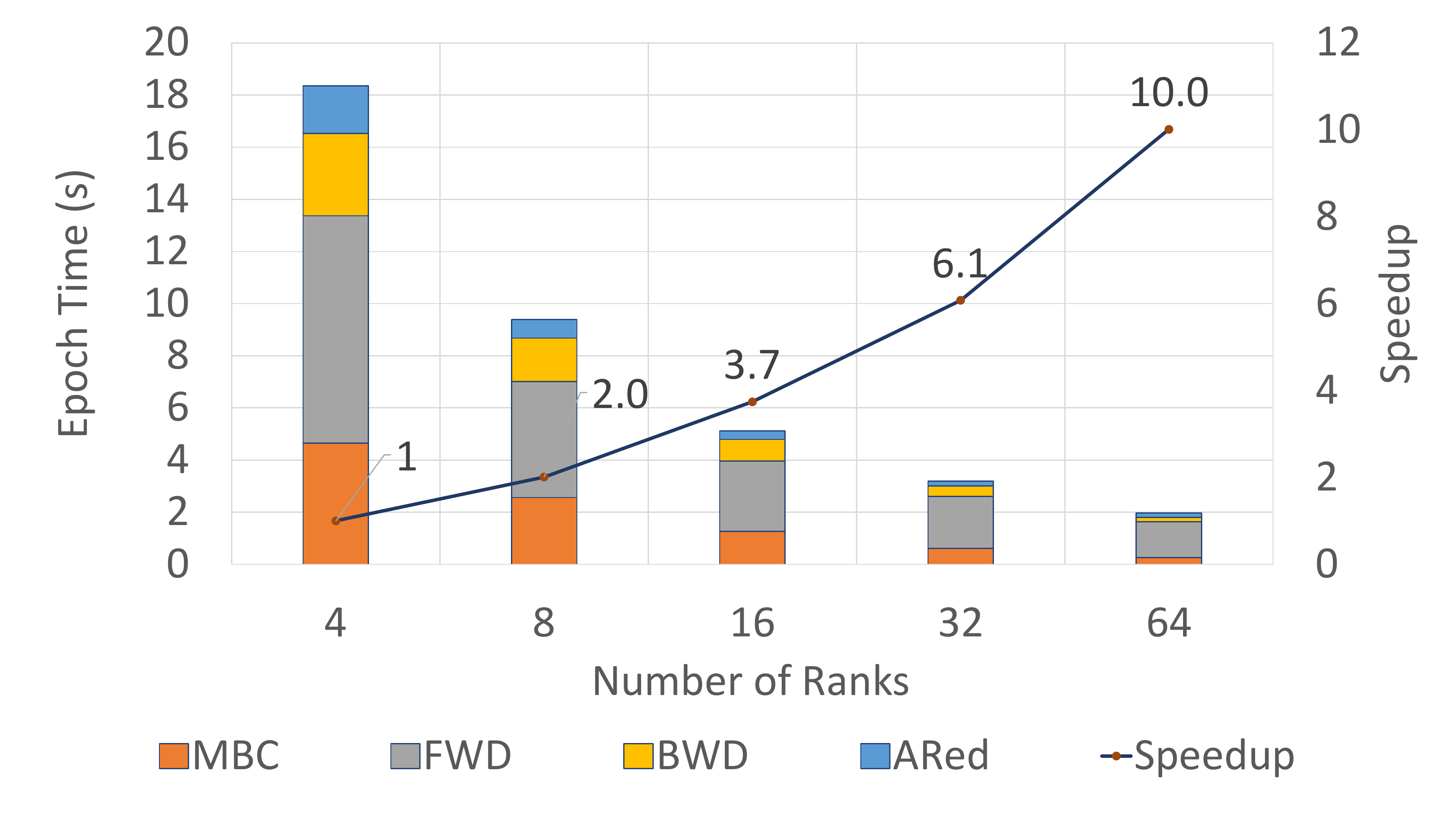}
    \caption*{OGBN-Papers100M}
    \label{fig:gsage-perf-2}
\end{subfigure}
\caption{Epoch time (in sec) and relative speedup for GraphSAGE from $2$ to $64$ ranks on OGBN datasets}
\label{fig:gsage-perf}
\end{figure*}

\begin{figure*}[!ht]
\begin{subfigure}{0.48\textwidth}
    \centering
    \includegraphics[width=\textwidth]{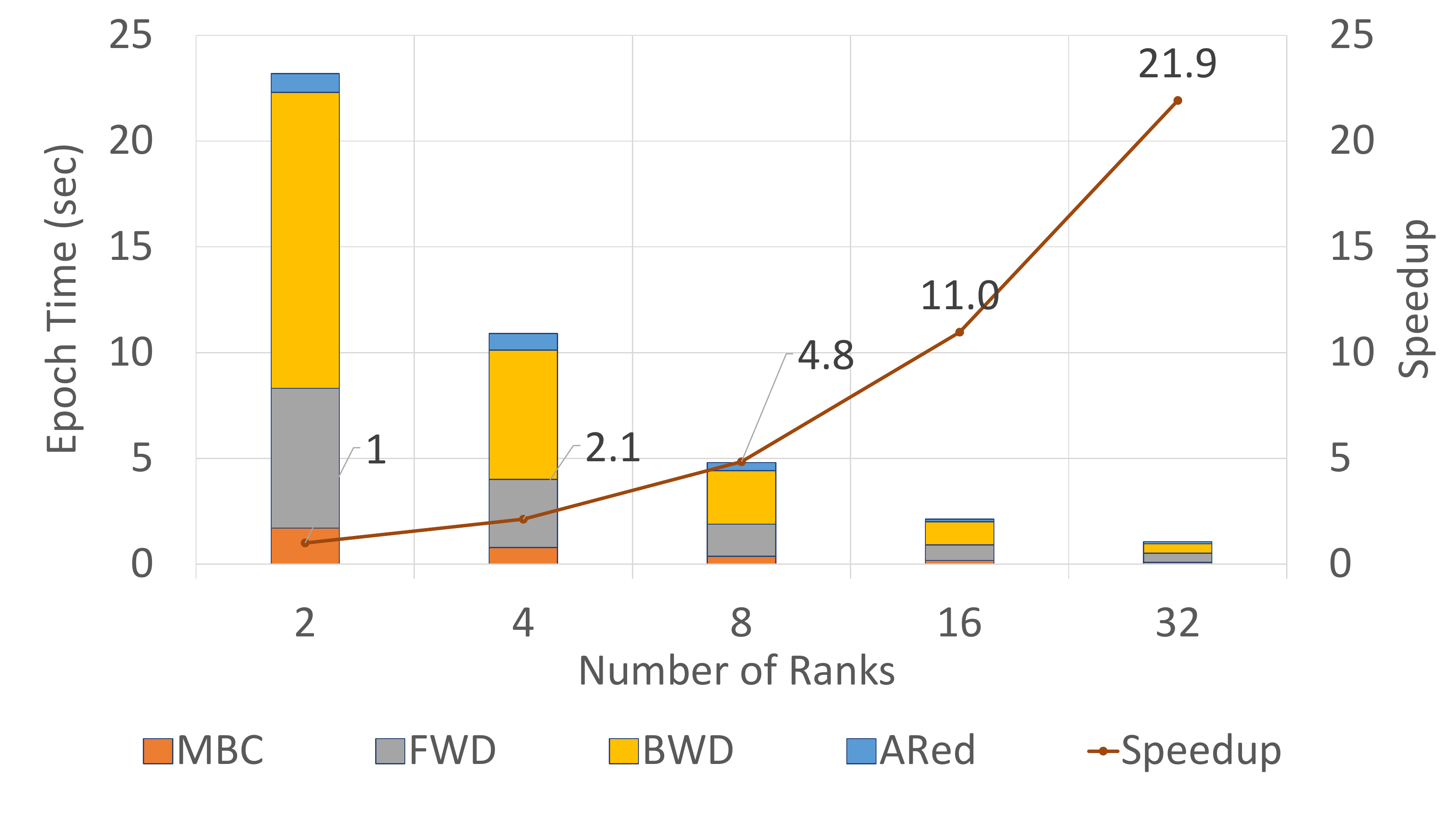}
    \caption*{OGBN-Products}
    \label{fig:gat-perf-1}
\end{subfigure}
\begin{subfigure}{0.48\textwidth}
    \centering
    \includegraphics[width=\textwidth]{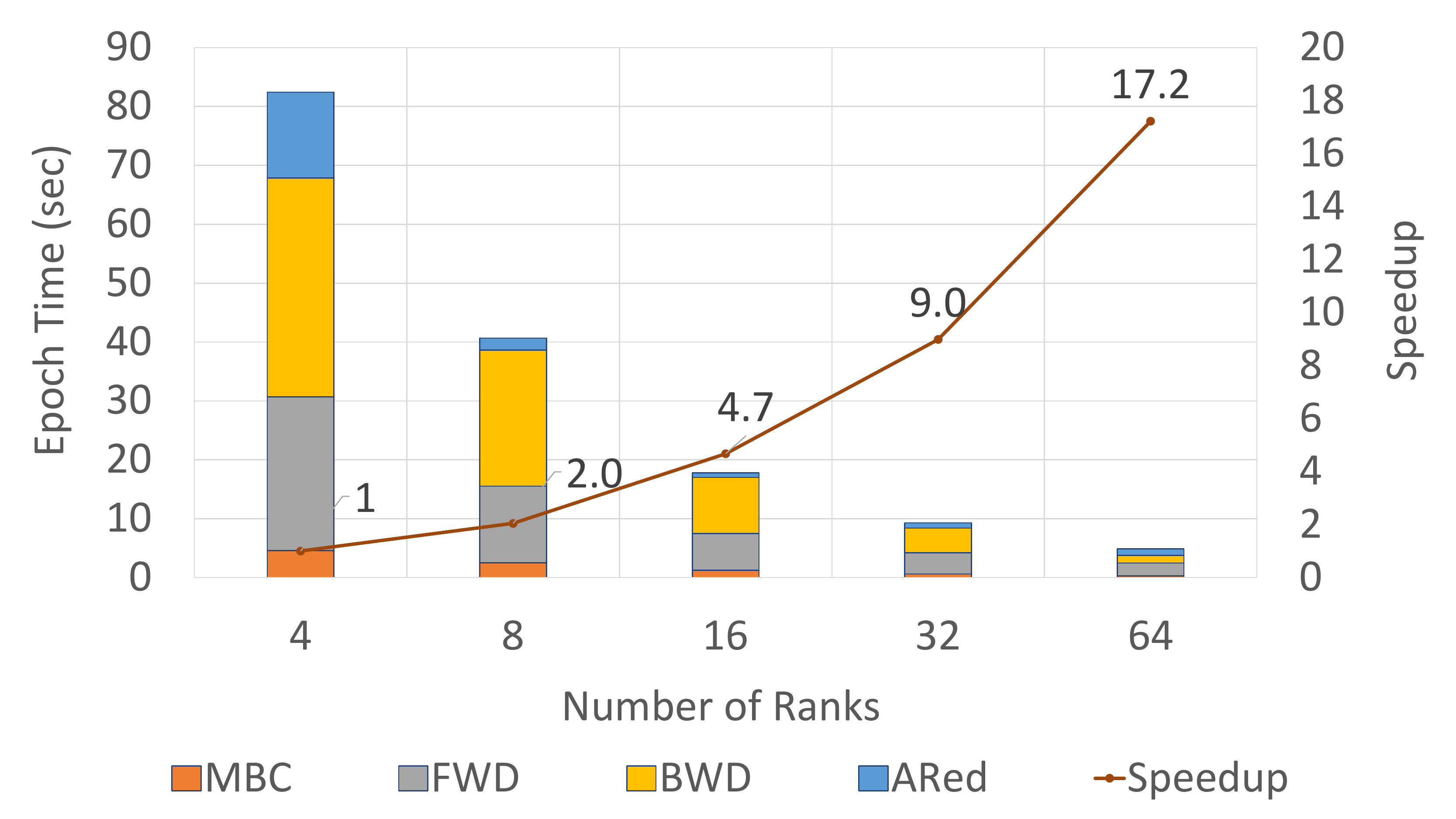}
    \caption*{OGBN-Papers100M}
    \label{fig:gat-perf-2}
\end{subfigure}
\vspace{-10pt}
\caption{Epoch time (in sec) and relative speedup for GAT from $2$ to $64$ ranks on OGBN datasets} 
%\zeroc{} epoch time is shown as horizontal dashed lines on each to illustrate the best achievable epoch time.}
\label{fig:gat-perf}
\end{figure*}

During model training, epoch time consists of the following components: minibatch creation time (MBC), forward pass time (FWD), backpropagation time (BWD), and MPI All-Reduce time (ARed) to aggregate model parameter gradients. FWD components, in addition to forward pass compute, also encompass remote aggregation time (pre-processing, communication and post-processing).
Figures~\ref{fig:gsage-perf} and ~\ref{fig:gat-perf} show the epoch time (including its components) and speedup of GraphSAGE and GAT models on OGBN datasets.

GraphSAGE on OGBN-Papers100M, epoch time consistently diminishes as we scale the number of ranks from $4$ -- $64$, with best epoch time of $2$ sec at $64$ ranks.
At $64$ ranks, GraphSAGE achieves best relative speedup (w.r.t. $4$ ranks) of $10\times$.
Among the components of epoch time, MBC and BWD scale perfectly linearly, while FWD and ARed run at 40\% and 69\% scaling efficiency from $4$ - - $64$ ranks, respectively. While \hec{} is able to completely hide the communication cost with a delay of $1$, pre- and post-processing time for communication starts dominating FWD with increase in scale. For example, with increasing number of ranks, more iterations over constant time  {\tt Map} (Algorithm~\ref{algo:push-model}) function leads to higher communication processing time. On OGBN-Products, GraphSAGE achieves $7.5\times$ speedup (w.r.t. $2$ ranks) on $16$ ranks.

For GAT, we see similar performance behaviour as GraphSAGE. On OGBN-Papers100M, it showcases the best epoch time of $4.9$ sec at $64$ ranks. %\he{} is xx\% - yy\% slower to \zeroc{}.
At $64$ ranks, GAT achieves best relative speedup (w.r.t $4$ ranks) of $17.2\times$. 

BWD dominates GAT epoch time, followed by compute part of FWD. Using \he{} leads to lowering of required compute. Consequently, we see greater than 100\% parallel efficiency as we scale from $4$ -- $64$ ranks. MBC and BWD scale perfectly linearly. FWD (due to pre- and post-processing for communication) and ARed run at 74\% and 85\% scaling efficiency, respectively, from $4$ -- $64$ ranks. Due to efficient BWD scaling, FWD time dominates at $64$ ranks.
On OGBN-Products, GAT shows $21.9\times$ speedup (w.r.t. $2$ ranks) on $32$ ranks.

On \hec{} utilization, we empirically observe that \hec{} hit-rate at $64$ ranks under chosen parameter settings ($cs=1M/layer$, $ls=2$, $nc=2000$ and $d=1$) is 71\%, 47\%, and 37\% at layers L$_0$, L$_1$, and L$_2$, respectively.

{\bf Load Imbalance.}
Multiple factors and their combinations lead to load imbalance in minibatch execution time. 
Uneven distribution of training examples (even after balancing attempts) among the partitions due to graph partition directly contributes to imbalance during training. For example, at $4$ ranks, lowest and highest minibatch counts across ranks are $264$ and $315$, respectively. Better distribution of training vertices during partitioning can alleviate the load imbalance problem.
Other, complex factors include number of halo vertices (or solid vertices) and cache hit-rate. %\revise{Randomization of halo node allocation per partition and selection per batch}
From $4$--$64$ ranks, we observe maximum load imbalance of 12\% and 8.7\% for GraphSAGE and GAT, respectively.

% \subsection{Large-Scale dataset: web-scale graph}
% Impact of various HEC params on convergence: cs, nc, ml

\subsection{Convergence}
%% TTC
% comparison of HE with 0c

We establish single-socket CPU test accuracy (by training for $20$ and $50$ epochs for OGBN-Products and OGBN-Papers100M) as target accuracy to assess the distributed training convergence (Table~\ref{tab:target-acc}). 

\begin{table}
\centering
\caption{Target test accuracy of GraphSAGE and GAT on OGBN datasets.}
\begin{tabular}{l|ccc}
\hline
     & \multicolumn{2}{c}{Test Accuracy (\%)} &  \\ \hline
Model/Dataset   & OGBN-Products      & OGBN-Papers100M &  \\  \hline
GraphSAGE       & $78.2$             &     $65$        &  \\
GAT             & $78.5$             &     $65$        & \\  \hline
\end{tabular}
\label{tab:target-acc}
\end{table}
We train the model till it achieves test accuracy within $1\%$ of the target accuracy (i.e., $target$ $accuracy - model$ $accuracy < 1\%$). 
In single-socket CPU runs, GraphSAGE converges at $7^{th}$ and $9^{th}$ epoch on OGBN-Products and OGBN-Papers100M, respectively; GAT converges at $6^{th}$ epoch for both the datasets. 
In our distributed experiments, we observe that GraphSAGE converges at the $30^{th}$ on OGBN-Products for $16$ ranks, and at the $15^{th}$ epoch on OGBN-Papers100M for $64$ ranks. Similarly, we observe that GAT converges at the $20^{th}$ on OGBN-Products for $32$ ranks, and at the $26^{th}$ epoch on OGBN-Papers100M for $64$ ranks.

%Give the target test accuracy (Table~\ref{tab:target-acc}) ....

%\subsection{Cache Analytics}
% HEC hit rate, HEC occupancy
% comms reduction/avoidance

\subsection{Comparison with DistDGL}
\begin{figure}[ht]
    \centering
     \includegraphics[width=0.48\textwidth]{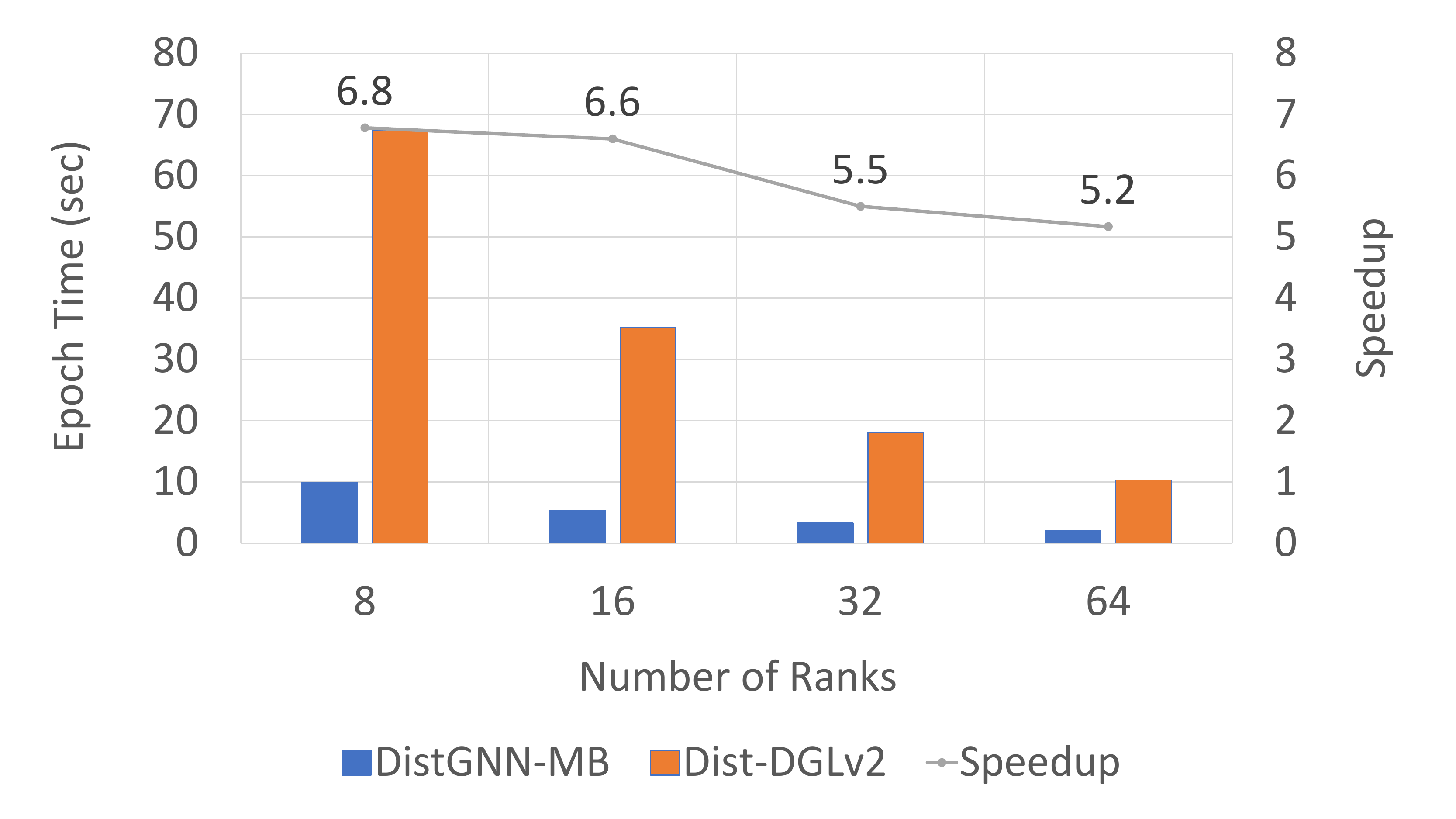}
    \caption{Performance comparison of \dgmb{} with DistDGL from $8$ to $64$ ranks using GraphSAGE mode on OGBN-Papers100M. We used $2$ sampler, $1$ trainer, and $1$ server for DistDGL experiments.  %OGBN-Products, OGBN-Papers100M.
    }
    \label{fig:comp-sota}
\end{figure}

% comparison on speed and time to converge
In this section, we compare the performance of \dgmb{} with state-of-the-art DistDGL~\cite{distdgl} for GraphSAGE. 
We ran GraphSAGE on DistDGL in our experimental setup following DistDGL instructions. For DistDGL, we used the exact same parameter settings as used in DistGNN-MB (described in section~\ref{sec-dataset}). Additionally, we used $2$ samplers, $1$ trainer, and $1$ server after limited tuning. 
Figure~\ref{fig:comp-sota} shows per epoch time comparison on OGBN datasets between \dgmb{} and DistDGL. \dgmb{} consistently performs better than DistDGL on both the datasets from $8$-- $64$ ranks. At $64$ ranks, \dgmb{} achieves speedup of $5.2\times$ per epoch over DistDGL. 

%% Comment on DistDGL scaling
%% Para on time to convergence

\section{Related Work}
\label{related}

Distributed GNN training is an important topic, ongoing research on batch creation, improving compute efficiency and mitigating communication overhead, improving model accuracy etc. In this section, we discuss specific efforts on training via {\em minibatch} sampling, including work on using historical embeddings, and compute-communication overlap in this setting (i.e., minibatch training). 

DistDGL~\cite{distdgl} is a distributed training architecture built on top of the Deep Graph Library (DGL); it employs a set of processes to perform distributed neighbor sampling and feature communication, a distributed key-value-store (KVStore) to hold vertex and edge features/embeddings, and a set of compute resources (processes/threads) to perform model training. It uses METIS~\cite{karypis1997metis} to partition the input graph using minimum edge-cut algorithm, but also to balance training example vertices as well as edges across machines.

Although Sancus~\cite{sancus} focuses on full-graph training on GPUs, we discuss their historical embedding technique that is "staleness-aware" and designed to reduce communication while maintaining accuracy. Sancus distributes both the adjacency matrix and node features tensor across GPUs; a "root" GPU performs a staleness-check and triggers node-feature-broadcast across all GPUs if embeddings become "too stale", but within staleness bounds all GPUs use embeddings they compute locally, to train the model. They discuss a theoretical proof that bounded staleness does not result in accuracy loss.

In~\cite{ramezani2022learn}, the authors propose a mechanism to ignore cut-edges that resulted in partitions, and applying on LocalSGD algorithm to compute model gradients during minibatch training in a distributed setting. At the end of every training iteration, each machine communicates it's local gradients to a global server that contains the full graph; the server {\em corrects} local gradients by  constructing a minibatch with full neighborhood, computing a stochastic gradient, and updating model parameters which it broadcasts back to all machines.

In~\cite{zeng2021166} and~\cite{graphsaint}, the authors propose {\em graph sampling} from the original un-partitioned graph, constructing a GNN on the sampled graph and training the model using each graph sample as a minibatch. They describe algorithms to eliminate the bias introduced by graph sampling and parallelization strategies to improve performance of this operation. Because they construct a full GNN on each sample, they do not need to communicate features during training. They also parallelize forward and backpropagation passes on each minibatch during training.

The Dorylus~\cite{dorylus} distributed system for training GNNs separates computation into "graph parallel" and "tensor parallel" tasks, executing the former on regular CPU instances and the latter on Lambda threads (low-cost, serverless threads) in an AWS environment. To ensure overlap between the two computation types, this work introduces {\em bounded pipeline asynchronous computation} to hide the latency incurred due to Lambda threads.

$P^3$~\cite{gandhi2021p3} uses a combination of model, data and pipelined parallelism to efficiently execute distributed GNN training on GPUs. It focuses on eliminating feature communication for the largest layer of the computational graph via model parallelism (i.e., each GPU owns part of the vertex features) and employing data parallelism at all other layers. To further mitigate GPU stalls due to lack of compute-communication overlap, $P^3$ implements pipelined parallelism. 

ByteGNN~\cite{bytegnn} is a distributed GNN training framework targeting efficient training performance on CPUs, with a primary focus on effectively scheduling minibatch creation tasks (including sampling, sub-graph creation and feature fetch) to overlap with model computation. To support the primary focus, this work also describes a graph partitioning algorithm that attempts to maintain data locality based on minibatch sampling data access patterns.

\section{Conclusions and Future Work}
\label{conclusions}

Large-scale graph structure learning by training GNN models is a very important technique that is growing in applicability and popularity, as many real-world data can be represented as graphs. To match the increasing scale of these graphs -- both in terms of size and quantity -- practitioners must develop techniques to improve GNN model training quality and speed. In this paper, we demonstrate the ability of a cluster of CPUs based on the 3rd generation Intel Xeon Scalable Processors to train two popular GNN models, GraphSAGE and GAT, to convergence at high-speed. To achieve these results, we developed a novel compute-communication overlap algorithm to reduce overhead at scale, a novel software-managed Historical Embedding Cache to reduce communications without accuracy impact, and performance optimizations to the \AGG{} primitive in \DGL{} and the \UPDATE{} primitive via a separate Pytorch C++ extension. As part of future work, we will expand benchmark coverage to demonstrate DistGNN-MB's wider applicability. We will also  demonstrate DistGNN-MB performance on the upcoming 4th generation Intel Xeon Scalable Processors with Advanced Matrix Instruction (AMX) and BFloat16 data-type support.

\bibliographystyle{ACM-Reference-Format}
\bibliography{distgnn}

\end{document}